\documentclass[twocolumn]{cinc}
\usepackage{mathtools, graphicx, xurl, hyperref}
\usepackage{tikz}
\usetikzlibrary{shapes,arrows,external,decorations.pathmorphing,backgrounds,positioning,fit,petri,calc,hobby,cd}
\usepackage{boldline,multirow}
\usepackage{booktabs} 
\usepackage{subcaption}
\usepackage{relsize}
\usepackage[ruled,vlined]{algorithm2e}

\title{Searching for Effective Neural Network Architectures for Heart Murmur Detection from Phonocardiogram}
\author{Hao Wen\textsuperscript{1},
Jingsu Kang\textsuperscript{2} \\ \ \\
\textsuperscript{1}LMIB and School of Mathematical Sciences, Beihang University, Beijing, China\\
\textsuperscript{2}Tianjin Medical University, Tianjin, China
}

\begin{document}
\maketitle



\begin{abstract}

Aim: The George B. Moody PhysioNet Challenge 2022 raised problems of heart murmur detection and related abnormal cardiac function identification from phonocardiograms (PCGs). This work describes the novel approaches developed by our team, Revenger, to solve these problems.

Methods: PCGs were resampled to 1000 Hz, then filtered with a Butterworth band-pass filter of order 3, cutoff frequencies 25 - 400 Hz, and z-score normalized. We used the multi-task learning (MTL) method via hard parameter sharing to train one neural network (NN) model for all the Challenge tasks. We performed neural architecture searching among a set of network backbones, including multi-branch convolutional neural networks (CNNs), SE-ResNets, TResNets, simplified wav2vec2, etc.

Based on a stratified splitting of the subjects, 20\% of the public data was left out as a validation set for model selection. The AdamW optimizer was adopted, along with the \texttt{OneCycle} scheduler, to optimize the model weights.


Results: Our murmur detection classifier received a weighted accuracy score of 0.736 (ranked 14th out of 40 teams) and a Challenge cost score of 12944 (ranked 19th out of 39 teams) on the hidden validation set.

Conclusion: We provided a practical solution to the problems of detecting heart murmurs and providing clinical diagnosis suggestions from PCGs.

\end{abstract}


\section{Introduction}
\label{sec:intro}

Heart murmur, defined as heart sounds produced by the turbulent blood flow through the heart, is a common clinical indicator in pediatric cardiology \cite{Masic_2018}. Accurate detection of heart murmurs and distinguishment between innocent murmurs and pathological murmurs help early clinical intervention of vital heart diseases such as congenital heart diseases, hence having a significant medical value.

Based on such motivations, the George B. Moody PhysioNet Challenge 2022 \cite{goldberger2000physionet,cinc2022} raised questions about detecting heart murmurs and identifying abnormal cardiac functions from phonocardiograms (PCGs), which are non-invasive heart sound recordings collected from multiple auscultation locations. In this paper, we present our methods of tackling these problems.

\section{Methods}
\label{sec:methods}



\subsection{Preprocess Pipeline}
\label{subsec:preproc}

After a careful study of spectral characteristics of heart murmurs from medical literature \cite{Noponen_2007}, and with reference to previous work \cite{Schmidt_2010}, we constructed the PCG signal preprocessing pipeline as follows:
\begin{itemize}
    \item Resampling to 1000 Hz;
    \item Butterworth bandpass filtering of order 3 and cutoff frequencies 25 - 400 Hz;
    \item Z-score normalization to zero mean and unit variance.
\end{itemize}

\subsection{Neural Network Backbones}
\label{subsec:backbone}

Inspired by the work of wav2vec2 \cite{baevski2020wav2vec}, and under the consideration of exploring and utilizing the powerfulness of pretraining models on larger databases, we adopted a shrunken \texttt{wav2vec2} as one of our neural network (NN) backbones. We used the time-domain signals, namely the PCG waveforms, as model input, rather than the derived time-frequency-domain signals, for example, the spectrograms. Since PCGs have significantly lower sampling rates than conventional human voice audio signals, we reduced the dimension (number of channels) of the `wav2vec2` model's encoder and its depth (number of hidden layers).


\begin{table}[!htp]
\centering
\begin{tabular}{r|l|l}
    \hline
    Backbone & \# Params & Input Type \\ \hline
    MultiBranch \cite{Kang_2022_cinc2021_iop} & 17.7M & waveforms \\
    SE-ResNet \cite{Kang_2022_cinc2021_iop} & 15.9M & waveforms  \\
    ResNet-NC \cite{ribeiro2020automatic} & 15.4M & waveforms  \\
    TResNetS \cite{Kang_2022_cinc2021_iop} & 41.0M & waveforms  \\
    TResNetF \cite{Kang_2022_cinc2021_iop} & 4.0M & waveforms  \\
    wav2vec2 \cite{baevski2020wav2vec} & 19.8M & waveforms \\ \hline
\end{tabular}
\caption{NN backbones tested for the Challenge tasks. wav2vec2 used the \texttt{transformers} implementation \texttt{Wav2Vec2Model} rather than the \texttt{torchaudio} implementation.}
\label{tab:nn_backbone}
\end{table}

Considering that PCGs share a similar physiological origin as electrocardiograms (ECGs), we further adjusted and tested several NN backbones that have proven effective in ECG problems, including MultiBranch CNN, SE-ResNet, TResNetS, TResNetF \cite{Kang_2022_cinc2021_iop}, and ResNet-NC \cite{ribeiro2020automatic} etc. We enlarged the kernel sizes of each convolution in these backbones by a factor of 2 (the ratio of the sampling rates).

\begin{figure}[!htp]
\centering
\includegraphics[width=\linewidth]{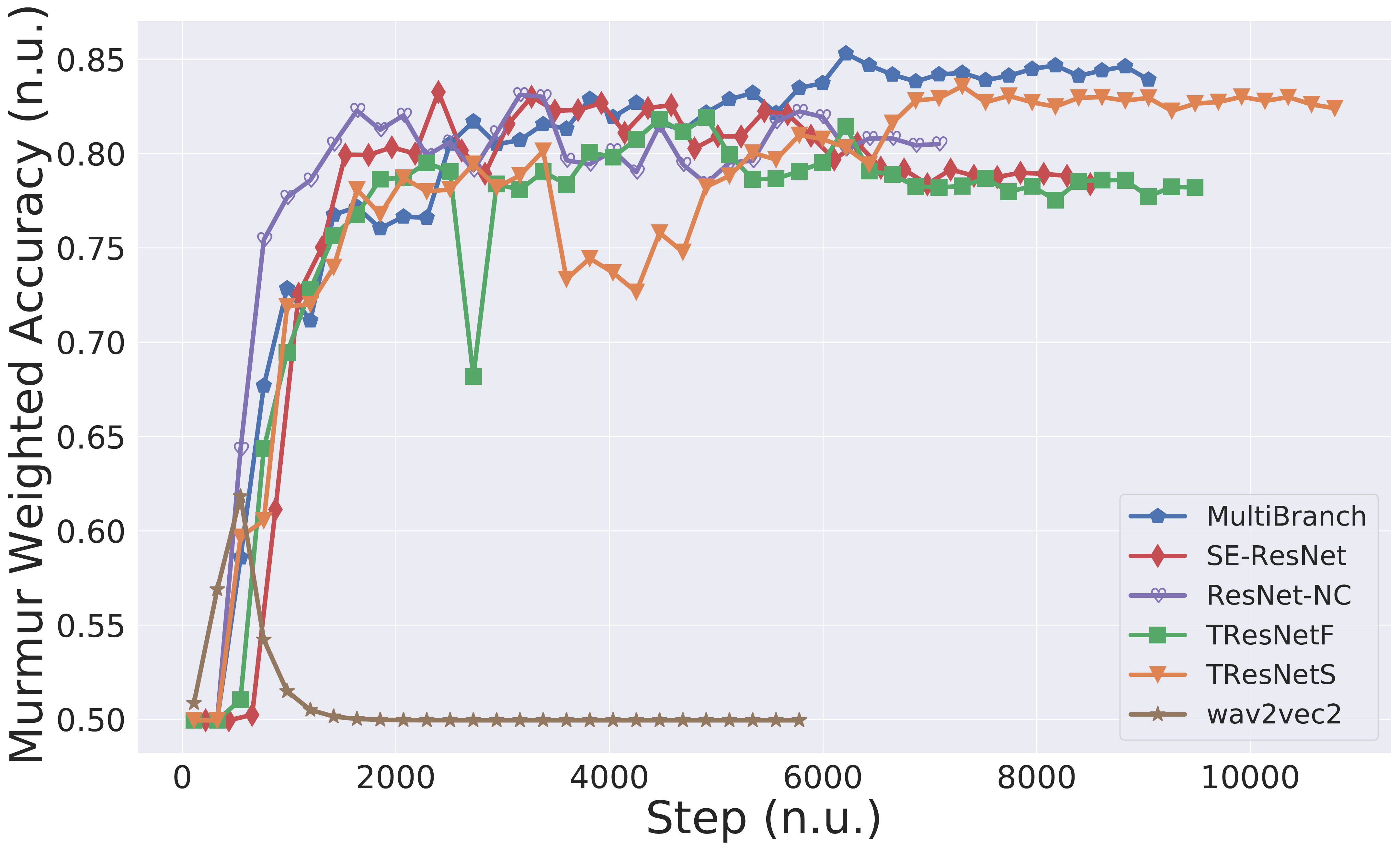}
\caption{Curves of weighted accuracies of murmur detection on the cross-validation set \ref{subsec:training} using 7 different NN backbones. The model heads, optimizers, loss functions, as well as other training setups, were kept the same.}
\label{fig:compare_nn}
\end{figure}

The efficacy of most of the NN backbones is validated via experiments as illustrated in Figure \ref{fig:compare_nn}. The learning process of the wav2vec2 model was interrupted at an early stage. The cause for this abnormal phenomenon is left for further studies.

\subsection{Multi-Task Learning}
\label{subsec:mtl}

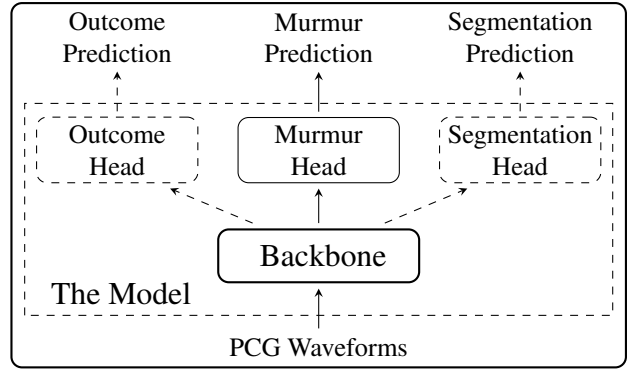
\begin{figure}[!htp]
\centering

\begin{tikzpicture}

\tikzstyle{block} = [rectangle, draw, text width = 5.5em, text centered, rounded corners, inner sep = 3pt, minimum height = 1.0em]
\tikzstyle{bigblock} = [rectangle, draw, text width = 7em, text centered, rounded corners, thick, inner sep = 3pt, minimum height = 2.0em]

\node (input) {PCG Waveforms};
\node [bigblock, above = 0.6 of input] (backbone) {\textbf{{\larger[2] Backbone}}};
\node [block, dashed, above left = 0.6 and 0.25 of backbone] (outcome) {Outcome Head};
\node [block, above = 0.6 and of backbone] (murmur) {Murmur Head};
\node [block, dashed, above right = 0.6 and 0.25 of backbone] (seg) {Segmentation Head};
\node [text width = 5.5em, text centered, above = 0.6 of outcome] (outcome_pred) {Outcome Prediction};
\node [text width = 5.5em, text centered, above = 0.6 of murmur] (murmur_pred) {Murmur Prediction};
\node [text width = 5.5em, text centered, above = 0.6 of seg] (seg_pred) {Segmentation Prediction};

\path[-stealth] (input) edge ([yshift = -2]backbone.south);
\path[-stealth, dashed] ([xshift = -25, yshift = 2]backbone.north) edge ([xshift = 20, yshift = -2]outcome.south);
\path[-stealth, dashed] ([xshift = 25, yshift = 2]backbone.north) edge ([xshift = -20, yshift = -2]seg.south);
\path[-stealth] ([yshift = 2]backbone.north) edge ([yshift = -2]murmur.south);
\path[-stealth, dashed] ([yshift = 2]outcome.north) edge (outcome_pred.south);
\path[-stealth, dashed] ([yshift = 2]seg.north) edge (seg_pred.south);
\path[-stealth] ([yshift = 2]murmur.north) edge (murmur_pred.south);

\draw[dashed] ([xshift = -35, yshift = -50]outcome.south) rectangle ([xshift = 35, yshift = 5]seg.north);
\node[below = 1.2 of outcome] {{\larger[1] The Model}};
\draw[rounded corners, thick] ([xshift = -40, yshift = -112]outcome_pred.south) rectangle ([xshift = 40, yshift = 0]seg_pred.north);

\end{tikzpicture}
\caption[]{The paradigm of multi-task learning (MTL) used in our team's approach. The dashed lines indicate optional model heads. The ``Outcome Head'' and the ``Murmur Head'' use pooled features from the ``Backbone'', while the ``Segmentation Head'' uses the unpooled features. The heads correspond to different tasks and share the same backbone.}
\label{fig:mtl_paradigm}

\end{figure}

The 2 Challenge tasks \cite{cinc2022} are per-patient classification tasks. It should be noted that the Challenge database \cite{Oliveira_2021_CirCor} provides per-recording annotations for the murmur detection task and heart sound segmentation annotations as well. We applied the multi-task learning (MTL) paradigm \cite{Caruana_1997_mtl} on each recording via hard parameter sharing. More precisely, we use one NN model for all the tasks. Each task has its specific model head, typically a stack of linear layers concatenated to the shared backbone as discussed in Section \ref{subsec:backbone}. Our MTL paradigm is illustrated in Figure \ref{fig:mtl_paradigm}.

\begin{figure*}
\centering
\begin{subfigure}[b]{0.49\linewidth}
    \centering
    \includegraphics[width=\textwidth]{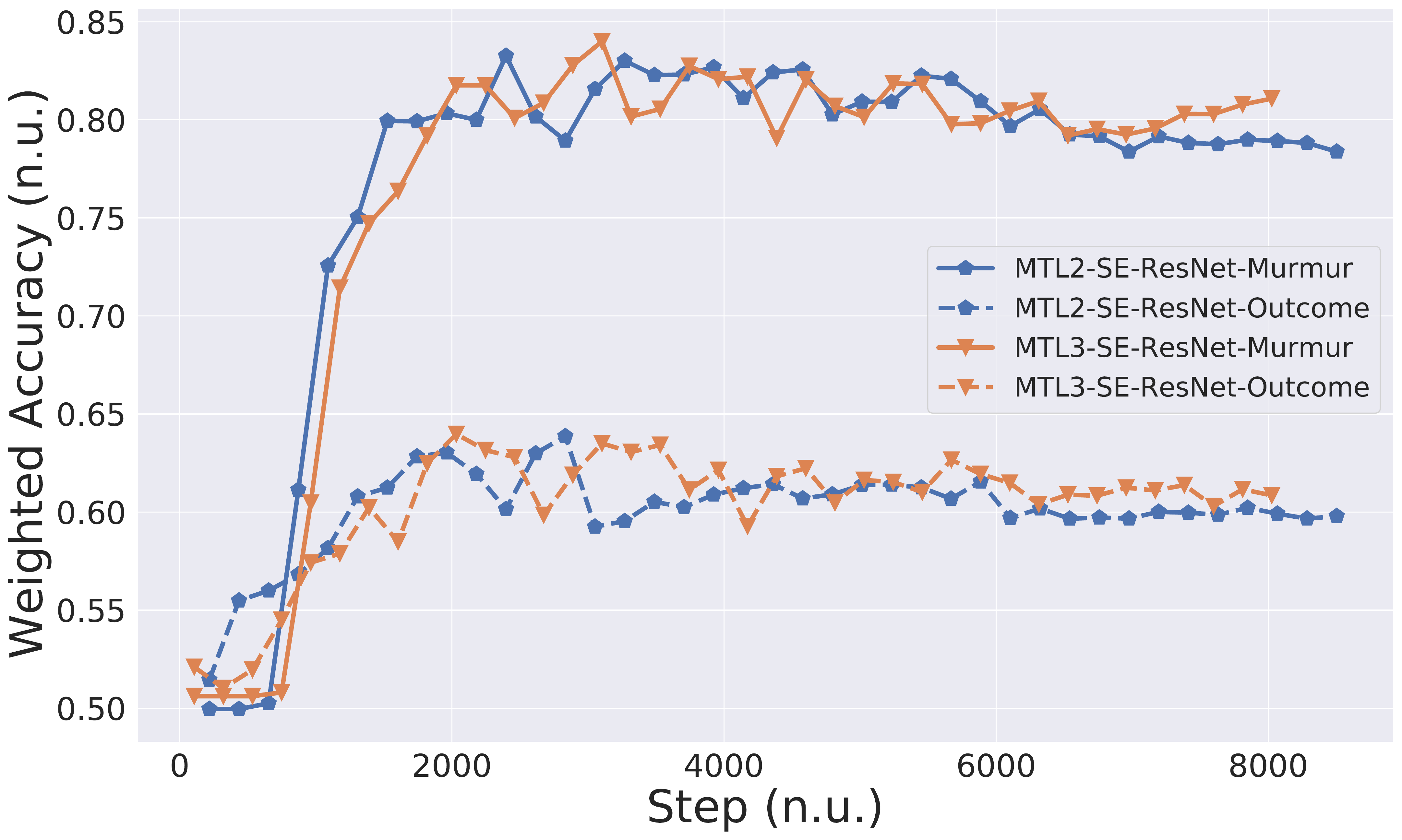}
    \caption[]
    {Experiments using SE-ResNet as the backbone.}
    \label{fig:se-resnet-clf-vs-mtl}
\end{subfigure}
\hfill
\begin{subfigure}[b]{0.49\linewidth}
    \centering
    \includegraphics[width=\textwidth]{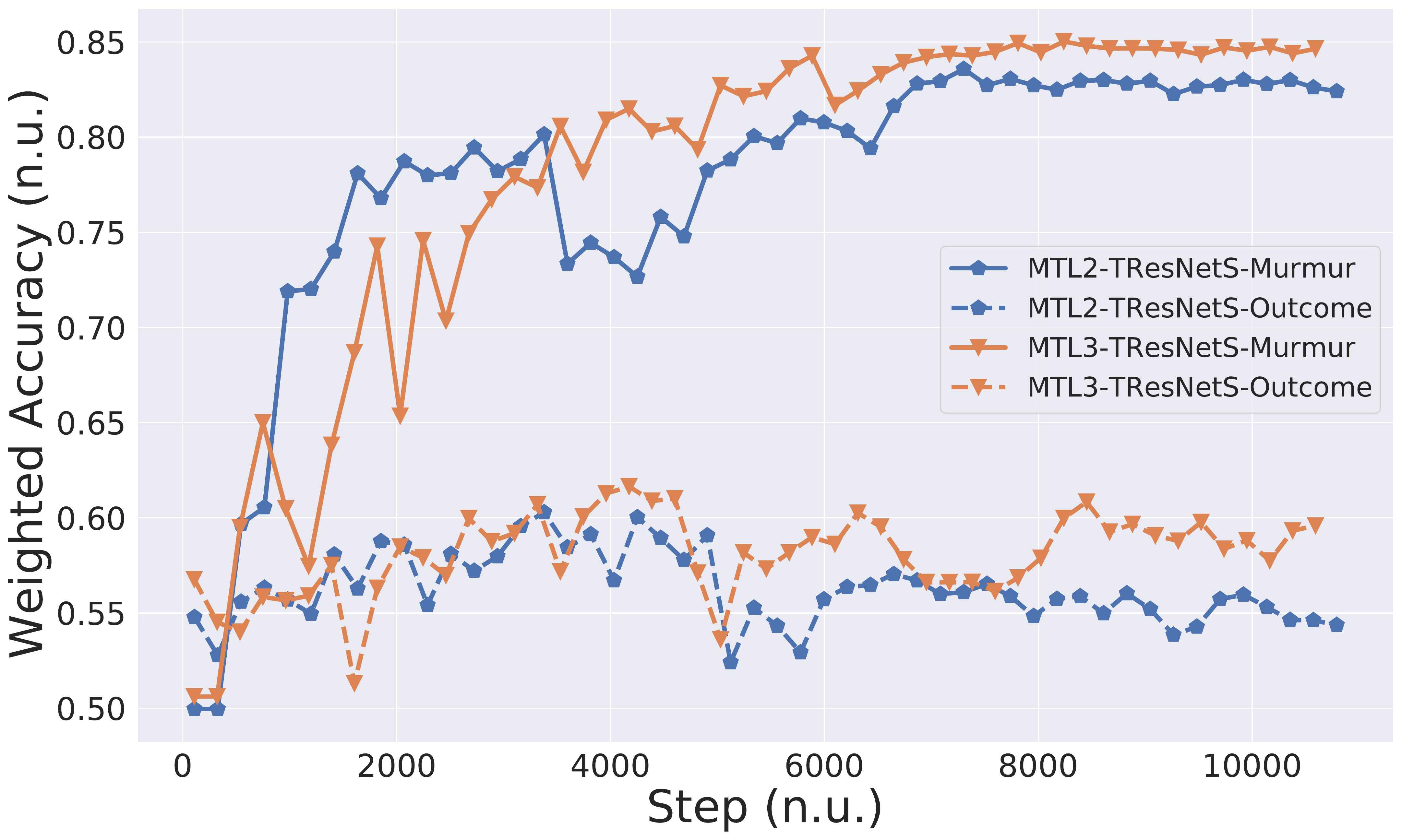}
    \caption[]
    {Experiments using TResNetS as the backbone}
    \label{fig:tresnets-clf-vs-mtl}
\end{subfigure}
\caption[]
{Experiments of the MTL method with 2 heads (for murmur classification and outcome classification) and with 3 heads (an additional head for heart sound segmentation) using 2 typical backbones.}
\label{fig:mtl_comparison}
\end{figure*}

As depicted in Figure \ref{fig:mtl_comparison}, experiments showed that models (with the same backbone) using an additional segmentation head (denoted as ``MTL3'') usually outperformed models with only two classification heads (denoted as ``MTL2'') for the Challenge tasks.

Our NN models produce per-recording predictions for the Challenge tasks. To obtain per-patient predictions, we used the simple greedy rule described in Algorithm \ref{alg:greedy}.

\begin{algorithm}
\SetInd{0.2em}{2em}
\uIf{at least on recording positive}{
    \textbf{Positive} for the patient\;
}
\uElseIf{all recording negative}{
    \textbf{Negative} for the patient\;
}
\Else(\tcp*[h]{for murmur detection only}){
    \textbf{Unknown} for the patient\;
}
\caption{The algorithm to obtain per-patient predictions}\label{alg:greedy}
\end{algorithm}

\subsection{Training Setups}
\label{subsec:training}

For algorithm development, we divided the publicly available part of the Challenge database into the training set and the cross-validation set with a ratio of 8:2. This split was stratified on the attributes ``Age'', ``Sex'', ``Pregnancy status'' and the prediction targets ``Murmur'', ``Outcome''.

The batch size was set at 32 for model training, with the maximum number of epochs set at 60. Model parameters were optimized using the AMSGrad variant of the AdamW optimizer \cite{adamw_amsgrad} along with the \texttt{OneCycle} scheduler \cite{smith2019one_cycle}. We froze the backbone from a specific epoch (usually 30), only updating the parameters of the task heads.

To alleviate overfitting on the training set, an early stopping callback was added. To further improve model transferability, we applied several types of augmentations to the batched training data stochastically:
\begin{itemize}
    \item adding coloured noises;
    \item polarity inversion (flipping).
\end{itemize}

\begin{figure}[!htp]
\centering
\includegraphics[width=\linewidth]{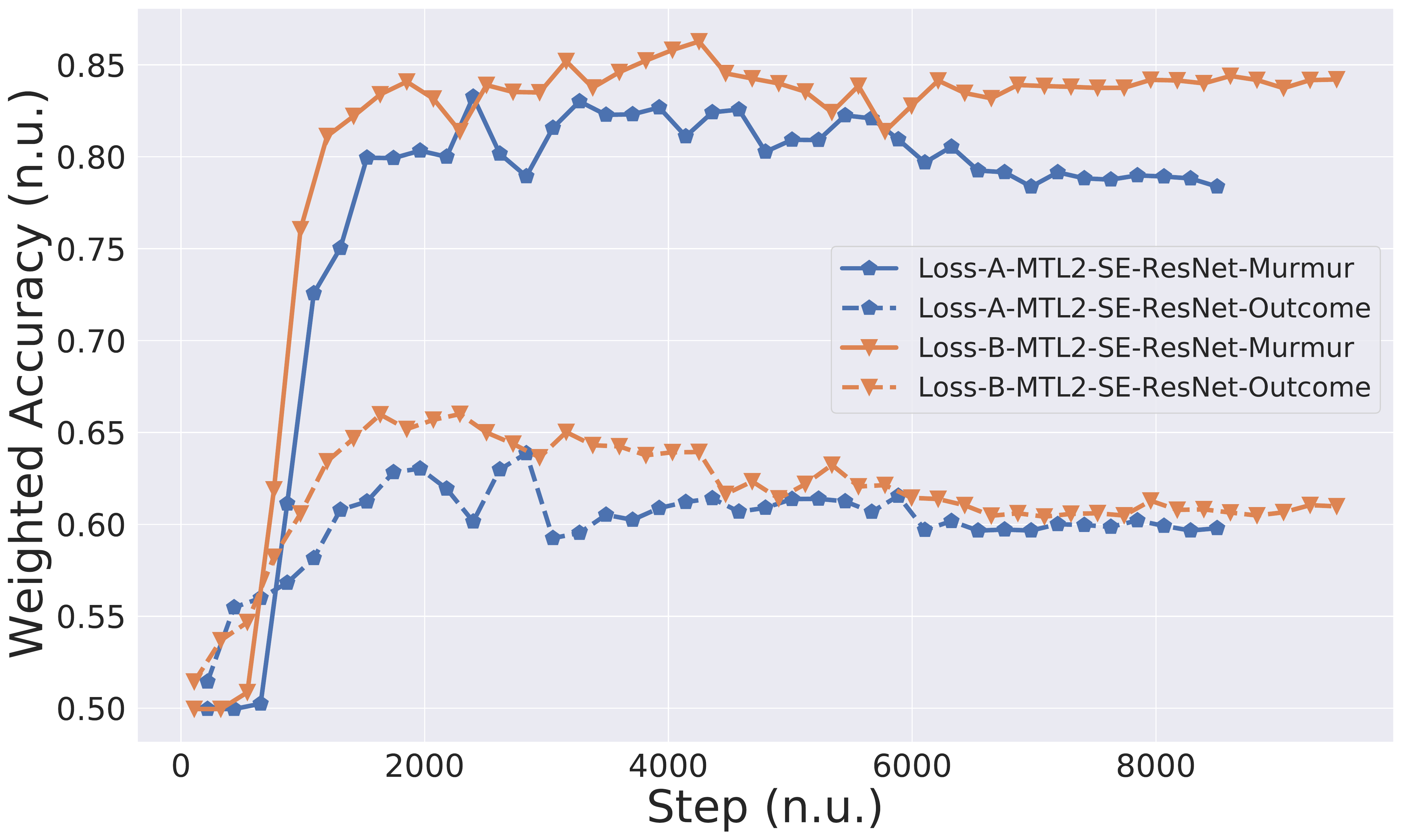}
\caption[]
{Experiments of comparison of the 2 loss functions (Loss-A for the asymmetric loss, Loss-B for the weighted BCE loss. The model with 2 classification heads and with SE-ResNet as the backbone was used.}
\label{fig:clf-se-resnet-lossA-vs-lossB}
\end{figure}

We experimented with two types of loss functions: the asymmetric loss, denoted ``Loss-A''; the weighted binary cross entropy (BCE), denoted ``Loss-B''. The weights were obtained from the weight matrix of the Challenge scoring functions \cite{cinc2022}. The superiority of Loss-B was observed, as was illustrated in Figure \ref{fig:clf-se-resnet-lossA-vs-lossB}.


\subsection{Demographic Features}
\label{subsec:demo_feat}

\begin{figure}[!htp]
\centering
\begin{subfigure}[t]{0.49\linewidth}
    \centering
    \includegraphics[width=\textwidth]{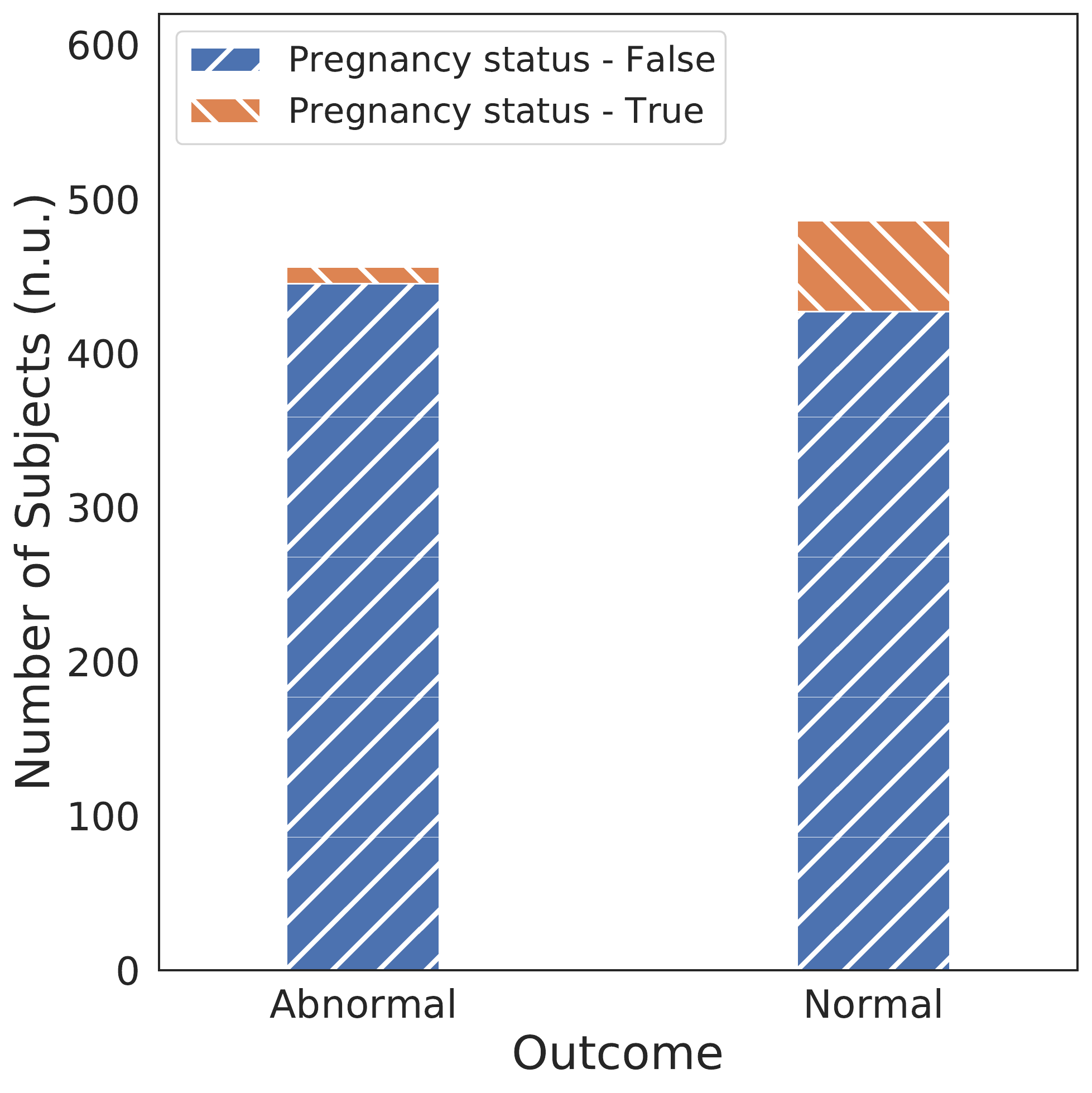}
    \caption[]
    {Dist. against pregnancy status.}
    \label{fig:outcome_pregnancy_status_corr}
\end{subfigure}
\hfill
\begin{subfigure}[t]{0.49\linewidth}
    \centering
    \includegraphics[width=\textwidth]{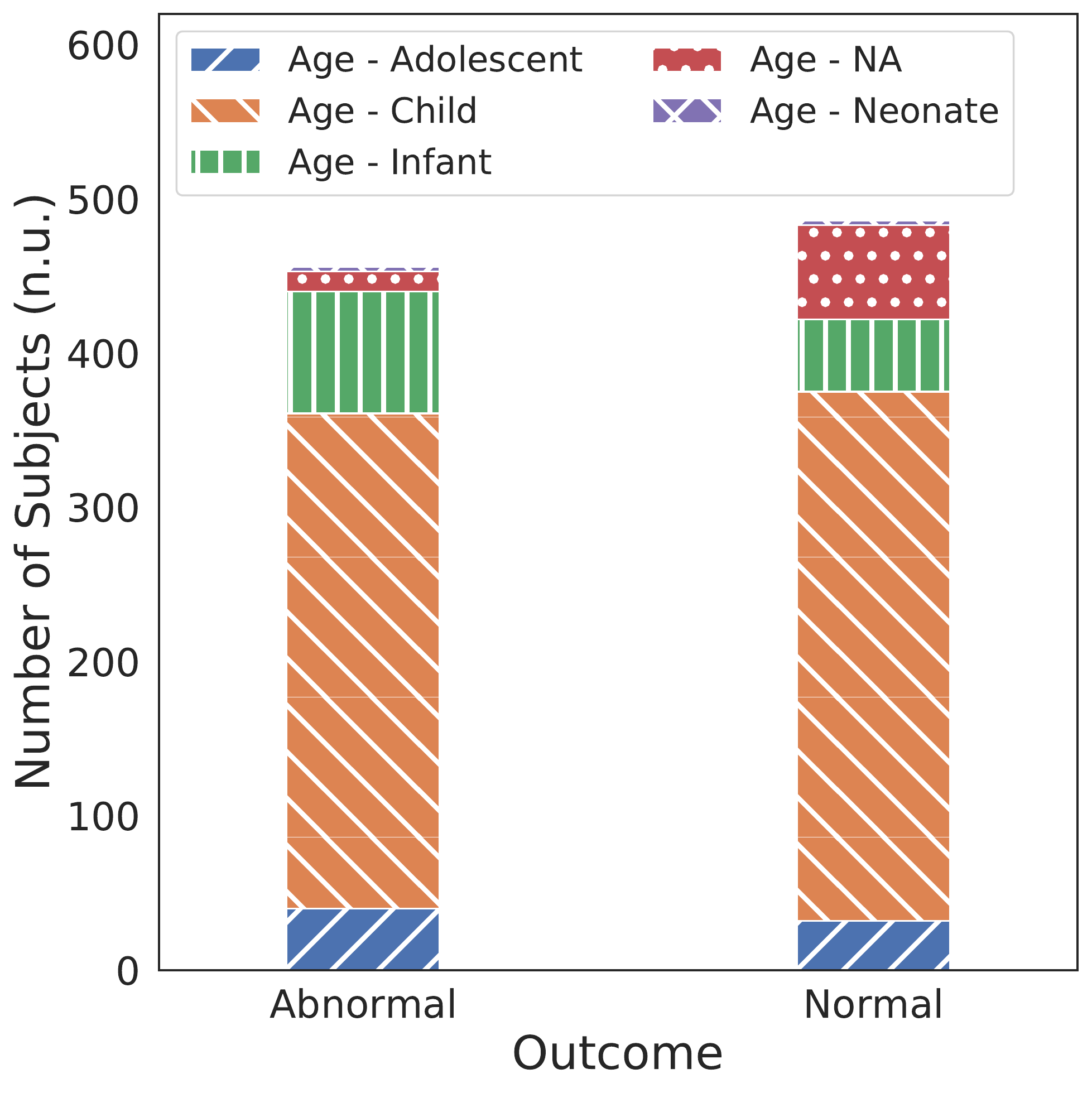}
    \caption[]
    {Dist. against age.}
    \label{fig:outcome_age_corr}
\end{subfigure}
\caption[]
{Distributions (Dist.) of the ``Outcome'' against 2 typical categorical demographic variables.}
\label{fig:outcome_corr}
\end{figure}

For the public data of the Challenge, some demographic features are strongly correlated with the prediction target ``Outcome'', as can be inferred from Figure \ref{fig:outcome_corr}. Experiments and official phase submissions showed that an auxiliary random forest classifier using these features and the murmur predictions improved the outcome scores (reduced the outcome cost). However, we did not use such auxiliary models in our final submission, since the distribution of these features might be completely different in the hidden data. Moreover, no supportive medical literature was found to support this point.


\section{Results}
\label{sec:results}

The Challenge scores (weighted accuracy for murmur detection and cost for clinical outcome identification) with an extra metric of weighted accuracy for clinical outcome identification on the train, cross-validation \ref{subsec:training}, the hidden validation, and the hidden test sets are gathered in Table \ref{tab:challenge_scores}. Scores on the former two sets are provided with mean and standard deviation over most of our offline experiments searching for the best NN architecture \ref{subsec:backbone}, \ref{subsec:mtl}, and loss functions \ref{subsec:training}.

\begin{table}[!htp]
\centering
\setlength\tabcolsep{2pt}
\begin{tabular}{@{\extracolsep{6pt}}c|c|c|c@{}}
    \hlineB{3.5}
    & \textbf{Murmur} & \multicolumn{2}{c}{\textbf{Outcome}} \\ \cline{3-4}
    & \textbf{wt. acc.} & \multicolumn{1}{c}{\textbf{cost}} & \multicolumn{1}{c}{wt. acc.} \\ \hline
    Training & 0.757 & 12244 & 0.770 \\
    Cross val. & $0.86\pm 0.01$ & $11341\pm 336$ & $0.79\pm 0.05$ \\ \hline
    Hidden val. & 0.678 & 10647 & 0.671 \\
    Hidden test & \textbf{0.736} & \textbf{12944} & 0.761 \\
    Ranking & \textbf{14 / 40} & \textbf{19 / 39} & 10 / 39 \\
    \hlineB{3.5}
\end{tabular}
\caption{Primary scores on the whole public training set, cross-validation set \ref{subsec:training} left out from the training data, hidden validation, and the hidden test set. ``wt. acc.'' is the abbreviation for weighted accuracy. Scores on the train and validation sets are of the form $mean \pm std. dev.$ calculated over all our offline experiments.}
\label{tab:challenge_scores}
\end{table}

\section{Discussion and Conclusions}
\label{sec:discu}

Our MTL paradigm proved effective for the problems of heart murmur detection and clinical outcome identification from PCGs in this study. The rankings of our team on the hidden validation and on the whole public training set were 20 / 40, 26 / 40 for murmur weighted accuracy, and 21 / 39, 32 / 39 for outcome cost respectively. These were all significantly lower than our rankings on the hidden test set as listed in Table \ref{tab:challenge_scores}. This phenomenon is not surprising, since the MLT paradigm has already shown to have the ability to improve generalizability via leveraging latent domain-specific knowledge inherited in the training data of related tasks \cite{Caruana_1997_mtl}. As for the problems that the Challenge raised, the additional segmentation head makes the shared representation (the common backbone) learn more general features and thus improves the performances for the original two classification tasks (heads).

The convolutional neural backbones also proved their effectiveness as has already shown in Figure \ref{fig:compare_nn}. Indeed, this figure exhibits only a small part of the architectures we had experimented with. However, there is still room for improvement, as compared to the top teams on the Challenge leaderboard.

One regret of this study is that the potential of using derived time-frequency-domain signals is not explored. Previous studies on various physiological signals have shown the powerfulness of neural networks combining the derived time-frequency-domain signals with the original time-domain signals.

Another weakness of this work is that we failed to use the wav2vec2 model for tackling the Challenge problems. One possible reason is that transformer-based models need to be trained on larger datasets, and perform worse on smaller datasets than CNNs. Using larger datasets to perform self-supervised pretraining for PCGs would be a direction for our future work.



\section*{Acknowledgments}

This work is supported by NSFC under grants No. 11625105, and 12131004.


\bibliographystyle{cinc}
\bibliography{references}

\begin{correspondence}
Jingsu Kang\\
No. 22, Qixiangtai Road, Heping District, Tianjin, China\\
kangjingsu@tmu.edu.cn,kjs890223@gmail.com
\end{correspondence}

\balance

\end{document}